\documentclass[lettersize,journal]{IEEEtran}
\usepackage{amsmath,amsfonts}
\usepackage{algorithmic}
\usepackage{algorithm}
\usepackage{array}
\usepackage[caption=false,font=normalsize,labelfont=sf,textfont=sf]{subfig}
\usepackage{textcomp}
\usepackage{stfloats}
\usepackage{url}
\usepackage{verbatim}
\usepackage{graphicx}
\usepackage{cite}
\usepackage{multirow}
\usepackage{colortbl}
\usepackage{xcolor}
\usepackage{booktabs}
\usepackage{pifont}        

\usepackage{booktabs}
\usepackage{tabularx}
\usepackage{caption}
\usepackage{siunitx} 

\RequirePackage{xspace}
\makeatletter
\DeclareRobustCommand\onedot{\futurelet\@let@token\@onedot}
\def\@onedot{\ifx\@let@token.\else.\null\fi\xspace}

\def\eg{\emph{e.g}\onedot}

\makeatother

\definecolor{rblue}{rgb}{0,0.5,1}
\definecolor{awesome}{rgb}{1.0, 0.13, 0.32}
\definecolor{hollywoodcerise}{rgb}{0.96, 0.0, 0.63}
\definecolor{lasallegreen}{rgb}{0.03, 0.47, 0.19}
\definecolor{hanpurple}{rgb}{0.32, 0.09, 0.98}
\definecolor{green(pigment)}{rgb}{0.0, 0.65, 0.31}

\usepackage[most]{tcolorbox} 

\newtcolorbox{bluequestion}{
  enhanced, breakable,
  colback=blue!5,        
  colframe=blue!30,      
  boxrule=0pt,           
  borderline west={6pt}{0pt}{blue!60}, 
  left=8pt, right=8pt, top=6pt, bottom=6pt, 
  arc=0pt                
}

\definecolor{cvprblue}{rgb}{0.21,0.49,0.74}
\definecolor{DeepPink}{rgb}{1,0.078,0.576}
\usepackage[pagebackref,breaklinks,colorlinks,allcolors=DeepPink]{hyperref}

\begin{document}

\title{{VL}2{Spike}: Spike-driven Distillation from VLMs for Low-Power Visual Perception in Embodied AI}


\author{Zinan Liu, Eric Zheng, Soumyaratna Debnath, Hao Shi, Ling Xiao, and Lin Wang$^{\dag}$~\IEEEmembership{Member,~IEEE}
\thanks{$^{\dag}$ Corresponding author.}
\thanks{Zinan Liu, Soumyaratna Debnath and Lin Wang are with the School of EEE, Nanyang Technological University (NTU), Singapore (email: zinan001@e.ntu.edu.sg; SOUMYARA004@e.ntu.edu.sg; linwang@ntu.edu.sg). }
\thanks{Eric Zheng is with Department of Computer Science, University of Toronto and Advanced Micro Devices, Inc., Canada (e-mail: Eric.Zheng@amd.com).}
\thanks{Hao Shi was a visiting student at the School of EEE, NTU, Singapore and is also with the State Key Laboratory of Extreme Photonics and Instrumentation, Zhejiang University, Hangzhou 310027, China (email:
haoshi@zju.edu.cn).}
\thanks{Ling Xiao is with the Faculty of Information Science and Technology, Hokkaido University, Japan (email:
ling@ist.hokudai.ac.jp).}}

\markboth{Journal of \LaTeX\ Class Files,~Vol.~14, No.~8, August~2021}%
{Shell \MakeLowercase{\textit{et al.}}: A Sample Article Using IEEEtran.cls for IEEE Journals}


\maketitle

\vspace{-15pt}
\begin{abstract}

Spiking neural networks (SNNs) are brain-inspired, event-driven models that compute with sparse spikes, which enables highly efficient visual perception in resource-constrained embodied AI models.
The emergence of Spiking-Transformer models with spike self-attention has substantially improved the learning capacity of pure SNNs. 
Although SNNs are energy efficient, their performance is still limited by the spike-based architecture and optimization challenges, as standard gradient descent rules cannot be directly applied.
Recently, vision-language models (VLMs) have shown rich multi-modal knowledge representation capabilities for visual perception. 
Thus, it is promising to leverage VLMs for better Spikformer training. 
To this end, we present \textbf{VL2Spike}, a novel spike-based knowledge distillation (KD) framework that bridges multi-modal knowledge from VLMs with compact Spikformer models. 
This design enhances the learning capacity of Spikformer models while preserving their energy-efficiency merits, thereby offering a practical pathway toward low-power robotic perception.
Our VL2Spike brings two key technical contributions. 
To align with spiking dynamics, we first propose 
\textit{spatial-temporal visual spike} (\textit{SVS}) 
distillation, which achieves \textbf{(1)} shared manifold alignment between VLM image features and spike tokens, and \textbf{(2)} warm-started temporal consistency on membrane potentials and spike rates.
We then design a novel 
\textit{spike prototype-guided linguistic} (\textit{SPL})
distillation strategy that aligns Spikformer's class prototypes and logits with promptable VLM text embeddings.
Extensive experiments show that VL2Spike achieves 
6.81\%
gain across three static datasets with only 15.7\% energy consumption.
It also exhibits strong generalization capacity on robotic visual place recognition (VPR) with a gain of
\textbf{6.63\%}, highlighting its potential for low-power perception in embodied AI.

\end{abstract}


\vspace{-10pt}
\section{Introduction}
\vspace{-5pt}
\label{sec:intro}

Spiking neural networks (SNNs) model neurons that communicate with time-stamped spikes rather than continuous activations, making computation event-driven and naturally sparse \cite{ghosh2009spiking, maass1997networks}. 
Such properties make SNNs particularly attractive for embodied AI and robotic perception systems, where visual understanding must be performed under strict energy, memory, and latency constraints.
Early theoretical foundations of SNNs were inspired by biophysical and phenomenological neuron models, such as Hodgkin-Huxley \cite{abbott2005model}, leaky integrate-and-fire \cite{gerstner2014neuronal}, and local learning rules such as STDP \cite{diehl2015unsupervised}. 
However, since gradient descent rules cannot be directly applied, practical training initially relied on artificial neural network (ANN)-to-SNN conversion \cite{bu2023optimal, wang2022signed}. 
It preserves firing rates while forfeiting precise temporal dynamics.


\begin{figure}[t!]
    \centering
    \includegraphics[width=1.0\linewidth]{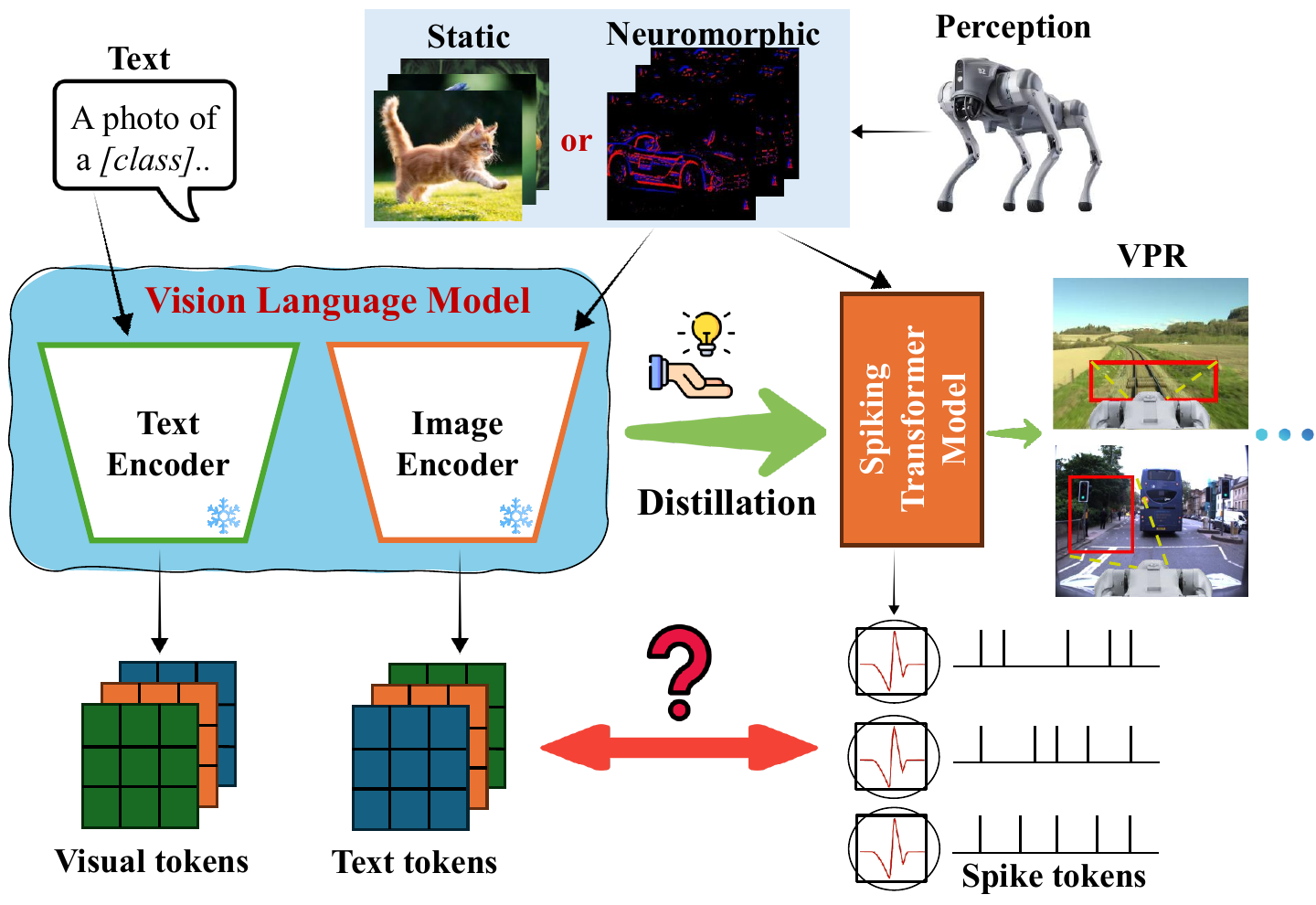} \\
    \vspace{-6pt}
    \caption{Our VL2Spike aims to bridge the multi-modal knowledge from VLMs and learn compact Spiking-Transformer models for low-power robotic visual perception.}
    \label{fig:teaser}
    \vspace{-15pt}
\end{figure}


More recent research on learning-based SNNs has combined spiking self-attention with rate-coding or temporal-coding and membrane potential dynamics to build Spiking-Transformer models  \cite{zhou2023spikingformer, zhou2022spikformer}. These models have pushed the pure SNNs to the best learning capacity.
Despite these gains, their performance is still limited by the spike-based architecture because gradient descent rules cannot be applied directly.

Recently, foundation models, particularly vision-language models (VLMs), \eg, \cite{cherti2023reproducible, liu2023visual}, couple powerful visual representations with promptable language priors. 
Through text prompts, these models expose class semantics that shape decision boundaries and enable zero-shot recognition in visual learning tasks \cite{awais2025foundation}. 
A frozen VLM exposes two complementary signals: visual features that capture category-discriminative structures across image tokens, and linguistic embeddings that encode class semantics through prompts. 
Therefore, it is promising to leverage them for better Spiking-Transformer training via knowledge distillation (KD) \cite{wang2021knowledge}.  
However, distilling multimodal knowledge is difficult when the student model has a spike-based structure. 
Unlike synchronous dense networks, Spiking-Transformer models process information with discrete spike tokens over time, operate under bio-inspired membrane dynamics, and favor sparse computation \cite{pfeiffer2018deep}. 
These properties enable excellent energy and latency trade-offs, 
which are critical for on-device robotic perception and low-power embodied agents,
but they also break the basic assumptions behind standard KD paradigms \cite{hinton2015distilling,wang2021knowledge}, which typically align intermediate features at a single time step. 
As a result, direct distillation from a VLM to spike-based models often produces fragile optimization or degraded efficiency (see Tab.~\ref{tab:comparison} and Fig.~\ref{fig:attention_map}). 


Essentially, two key challenges exist in multimodal grounding that respects spiking dynamics. 
We show that transferring only one signal is insufficient (see Tab.~\ref{tab:loss_ablation}). 
This is because visual alignment alone can converge to ambiguous decision boundaries (challenge 1), while text-only distillation can underconstrain spatial structure and be sensitive to prompt phrasing (challenge 2). 
To address these issues, we propose \textbf{VL2Spike}, a novel spike-based distillation framework that marries multi-modal knowledge from VLMs with compact Spikformer models. 
Our VL2Spike enhances the learning capacity of Spikformer models while preserving their energy-efficiency merits. 
This makes VL2Spike a practical step toward efficient multimodal perception for embodied AI that maintain semantic awareness under limited computational budgets.
VL2Spike is pluggable, thus it can be flexibly extended to neuromorphic data inputs, \eg, \cite{davies2018loihi}, or conventional SNN student models, \eg, \cite{deng2022temporal}. 
Our method brings contributions through two complementary objectives, as depicted in Fig.~\ref{fig:teaser}.

Specifically,  to address the first challenge, we propose a \textit{spatial-temporal visual spike }(\textit{SVS}) KD loss (Sec.~\ref{sec:visual_kd}). It projects VLM-based image features and student token features into a shared manifold, alignment is computed at corresponding patch locations and aggregated across time so that spatial topology is preserved while remaining efficient for few steps. To respect membrane dynamics, we then regularize temporal behavior with a consistency term on membrane potentials and spike rates that encourages smooth evolution across adjacent time steps and discourages gratuitous firing, thereby preserving sparsity. To tackle the second challenge, we propose a \textit{spike prototype-guided linguistic }(\textit{SPL}) KD loss (Sec.~\ref{sec:lingual_kd}). It injects the semantics by aligning the student’s class prototypes to prompt derived text embeddings from the teacher. The prototypes are formed as an exponential moving average of class conditioned features and the current logits.  This stabilizes decision boundaries and improves calibration.

\noindent \textbf{Our main contributions} are summarized as: (\textbf{I}) We explore the first attempt by exploring a spike-based distillation pipeline that transfers multimodal knowledge from a frozen VLM into compact Spiking-Transformer students, enhancing learning capacity while preserving the \textbf{energy efficiency} of SNNs. (\textbf{II}) We align image features of VLM with student's spike tokens by projecting both of them into a shared manifold at corresponding locations, preserving spatial structure for temporal consistency. (\textbf{III}) We inject language semantics by aligning student class prototypes and logits with prompt derived text embeddings from the VLM, stabilizing decision boundaries and improving calibration. (\textbf{IV}) We yield consistent accuracy gains up to \textbf{~6.81\%} on static benchmarks and transfer well to neuromorphic datasets and VPR tasks, also strong generalization capacity for conventional SNN student models.  

\vspace{-8pt}
\section{Related Work}
\vspace{-3pt}
\label{sec:formatting}

\noindent\textbf{Spiking Neural Networks (SNNs)} are widely regarded as the `third generation' of neural models, where computation and communication rely on discrete spikes rather than continuous activations~\cite{maass1997networks}. This event-driven paradigm offers low latency and energy efficient inference. However, the non-differentiability of spikes necessitates surrogate gradient methods to enable end-to-end training~\cite{neftci2019surrogate}. Building on this, hybrid convolution spiking architectures reduce computation while preserving accuracy~\cite{xu2022hierarchical}, and frequency domain or compact-time formulations further decrease the required simulation steps without sacrificing performance~\cite{yu2025fsta}. Trainable depth has been extended through spike-aware residual designs and element-wise calibration, reaching ImageNet-scale and $100+$ layer SNNs~\cite{fang2021deep,hu2024advancing}. Recently, Transformer style SNNs with spiking self-attention and improved tokenization, normalization, pooling, and meta-learning achieve the best learning capacity on both neuromorphic and static datasets~\cite{zhou2022spikformer, zhou2023spikingformer, yao2024spike}. To further push the efficiency and scalability of spiking transformers, recent advancements have explored token pruning techniques to reduce computational overhead in resource-constrained scenarios~\cite{wei2026tpspikformer}, and fully spike-driven transformers that integrate multi-view learning for sequential data processing~\cite{wang2025spikcommander}.


\noindent \textbf{ANN-to-SNN distillation.} KD has become a strong alternative to pure ANN-to-SNN weight conversion, transferring supervision from high-accuracy ANNs to efficient SNN students \cite{yang2025efficient, yu2025efficient, qiu2024self, zuo2024self}. Early works distilled logits and spike statistics. {KDSNN} combined logits- and feature-based guidance to reduce the accuracy gap induced by binary spikes and temporal dynamics \cite{Xu2023KDSNN}, and {LaSNN} introduced a layer-wise ANN-to-SNN scheme with attention-based feature alignment \cite{Hong2023LaSNN}. More recently, {BKDSNN} restored blurred teacher features to better match ANN intermediate representations \cite{Xu2024BKDSNN}. Designing feature-level losses for heterogeneous ANN to SNN teachers and students remains non-trivial because the student state is spike-driven and temporally unfolded. Prior KD either aligns high-level logits \cite{Xu2023KDSNN} or transfers intermediate feature maps with layer-wise objectives \cite{Hong2023LaSNN}, and directly matching continuous ANN features to discrete rate- or time-coded SNN states can cause gradient mismatch and optimization instability. A recent study further highlights this intrinsic discrepancy between continuous ANN distributions and sparse SNN outputs, proposing noise-smoothed logit distillation and saliency-scaled activation mapping to mitigate the representation gap~\cite{liu2025closer}. For Spikformer, where attention tokens and temporal spikes interact, this further motivates mediator spaces and token- or relation-aware penalties. However, existing ANN-to-SNN KD losses are tailored to frame-aligned feature maps or logits between homogeneous vision backbones, and therefore incompatible with our VLM setting where the teacher provides language-conditioned token relations instead of temporally unfolded spike states.


\noindent \textbf{KD from Foundation Models.}
Distilling multimodal knowledge from VLMs introduces a second source of heterogeneity, aligning spiking visual features with continuous text embeddings. Recent VLM to vision distillation, \eg, \cite{Jang2025VL2Lite, zhang2024vlm}, demonstrates that injecting language-informed supervision into lightweight vision-based students can significantly improve recognition. Yet, mapping SNN representations
(\eg, membrane potentials and spike rates)
into text-encoder embedding spaces remains notably underexplored. Effective cross-modal transfer likely requires (i) a shared mediator space that normalizes SNN temporal states into modality-agnostic tokens, (ii) relational constraints that respect token-token structure, and (iii) objectives that preserve SNN efficiency. 


\noindent\textbf{Position of Our Work.} 
Differently, we propose \textbf{VL2Spike}, the {first} spike-based distillation framework that can marry multi-modal knowledge from VLMs with compact Spikformer models. Our VL2Spike subtly enhances the learning capacities of Spikformer models while balancing their energy efficiency merits. Our VL2Spike is pluggable, so it can be flexibly extended to neuromorphic inputs or conventional SNN student models. Technically, to align with the spiking dynamics, we first propose \textit{spatial-temporal visual spike distillation}. We also propose a novel \textit{spike prototype-guided linguistic} distillation that aligns Spikformer's class prototypes and logits to promptable VLM text embeddings.

\vspace{-6pt}
\section{Methodology}

\subsection{Overview and Problem Definition}
\begin{figure*}[h!]
    \centering
    \vspace{-15pt}
    \includegraphics[width=0.95\linewidth]{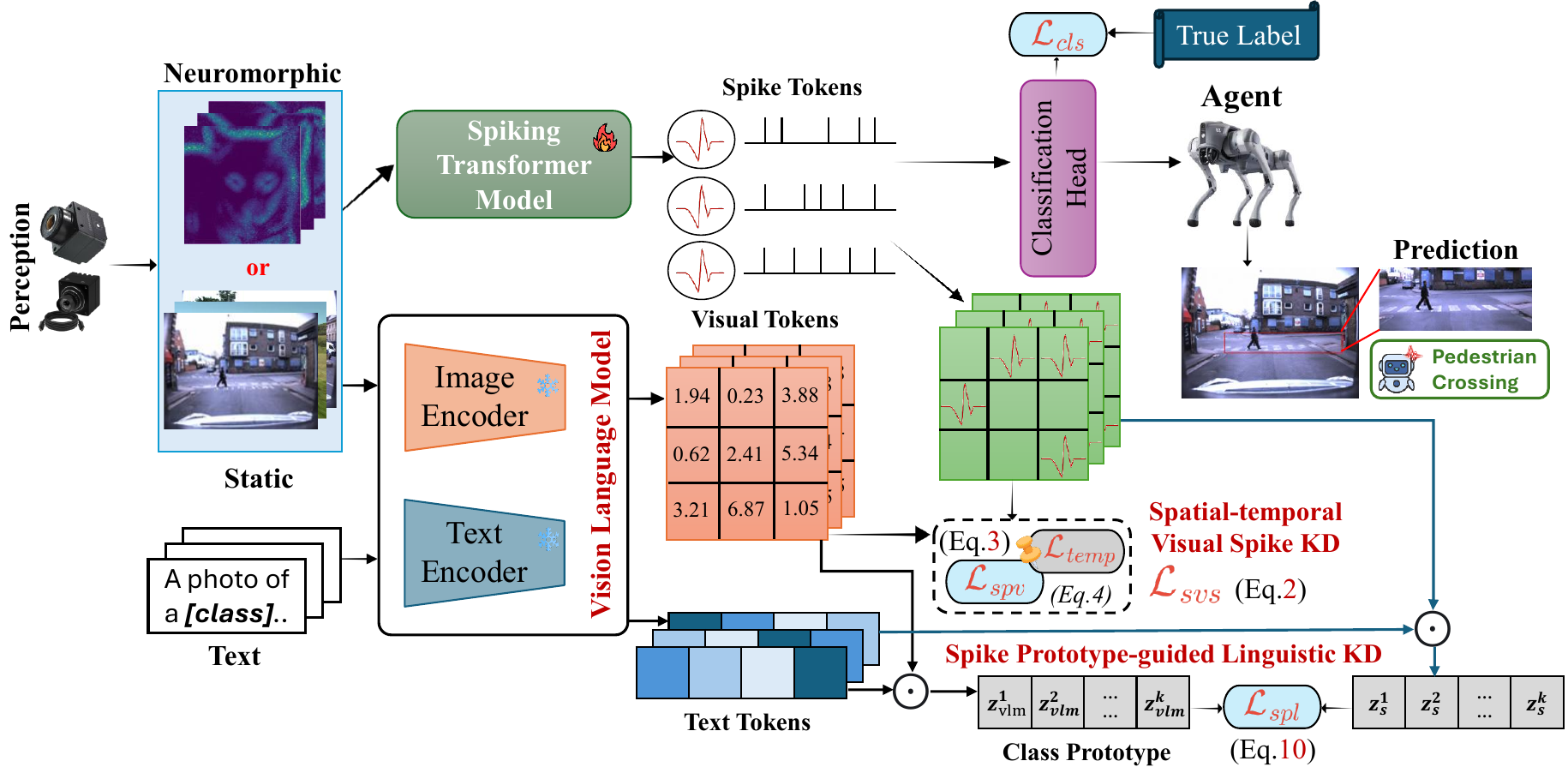}
    \vspace{-8pt}
    \caption{\textbf{An overview of our proposed VL2Spike framework that consists of two key objectives.} The \textit{spatial-temporal visual spike} (\textit{SVS}) distillation (Sec.~\ref{sec:visual_kd}) aims to achieve {(1)} shared manifold alignment between VLM image features and spike tokens, and {(2)} warm-started temporal consistency on the membrane potentials and spike rates. The \textit{spike prototype-guided linguistic} (\textit{SPL}) distillation (Sec. \ref{sec:lingual_kd}) that aligns Spikformer's class prototypes and logits to promptable VLM text embeddings. $\odot$ represents element-wise multiplication.}
    \label{fig:kd_spikformer}
    \vspace{-17pt}
\end{figure*}

\noindent \textbf{Overview.} Our distillation framework transfers both visual and linguistic knowledge from a frozen VLM into a Spikformer student as shown in Fig.~\ref{fig:kd_spikformer}. To tailor the student to the classification task, we use a standard cross-entropy objective on the Spikformer outputs. Given an input image $x$ with ground truth label $y \in \{1,\dots,N_{\text{cls}}\}$, the Spikformer student produces per class logits from its spiking tokens. We aggregate temporal responses over $T$ time steps and apply a linear classification head:
{\setlength\abovedisplayskip{2pt}
\setlength\belowdisplayskip{2pt}
\begin{equation}
\small 
\bar{\mathbf{z}}_s(x) \;=\; \frac{1}{T}\sum_{t=1}^{T}\mathbf{z}^{(t)}_s(x),
\qquad
\boldsymbol{\ell}_s(x) \;=\; \text{head}\!\left(\bar{\mathbf{z}}_s(x)\right),
\end{equation}}
where $\mathbf{z}^{(t)}_s(x)$ denotes the student feature at time step $t$. Class probabilities are calculated using softmax function $p_s(c\mid x)=\mathrm{softmax}(\boldsymbol{\ell}_s(x))_c$.

The core idea is to let the VLM act as a fixed semantic scaffold while the spiking student learns (i) teacher like visual relations among tokens and (ii) language-grounded decision boundaries, all while adhering to spiking dynamics and remaining efficient at low time steps.

We begin by aligning the student’s spike tokens with the VLM’s visual tokens through a structure-preserving visual loss that matches pairwise token relations, which is a robust Huber discrepancy, and a temporal-consistency regularizer that stabilizes discrete spikes against a smooth membrane potential trajectory (Sec.~\ref{sec:visual_kd}). To inject language, we perform linguistic knowledge distillation by prompting the frozen VLM’s text encoder to produce class prototypes and then aligning the student’s spike features to these prototypes via temperature scaled distributions (Sec.~\ref{sec:lingual_kd}). In parallel, a standard classification task objective optimizes recognition accuracy on labels. The result is a spiking student that inherits rich visual-lingual priors from a single VLM model, achieving higher accuracy, without sacrificing the event-driven efficiency of SNNs.

\vspace{-10pt}
\subsection{Spatial-Temporal Visual Spike Distillation}
\label{sec:visual_kd}
\noindent \textbf{Insights.} Distilling from a continuous, frame-based VLM into a discrete time spiking student is non-trivial since the two models differ in signal representation and temporal dynamics. The VLM features are valued real and temporally collapsed, whereas the SNNs communicate through discontinuous spikes that emerge from membrane potentials over a few time steps. The naive matching of the feature or logit ignores the spatial relations among the tokens that encode the layouts and boundaries of the objects, and neglects the temporal credit assignment that shapes spike timing.

Our Spatial-Temporal Visual Spike (SVS) distillation enforces a shared manifold alignment between VLM image features and student spike tokens by matching the pairwise token relations with a structure preserving visual loss that is robust to outliers (illustrated in Fig.~\ref{fig:kd_spikformer}). This preserves local geometry and global layout so that the student internalizes teacher like visual relations among tokens rather than regressing to per-token averages. To stabilize the discrete signaling process, SVS adds a warm started temporal consistency regularizer on membrane potentials and spike rates, encouraging smooth evolution across early time steps without collapsing informative spikes.
To transfer visual knowledge from the frozen VLM into the Spikformer student, we match the relational structure of the features and regularize the student’s spiking dynamics. The overall visual loss is:
{\setlength\abovedisplayskip{2pt}
\setlength\belowdisplayskip{2pt}
\begin{equation}
\label{eq:l_vis}
\mathcal{L}_{\mathrm{svs}} 
= \lambda_{\mathrm{spv}}\,\mathcal{L}_{\mathrm{spv}} + \lambda_{\mathrm{temp}}\,\mathcal{L}_{\mathrm{temp}}.
\end{equation}}
\noindent Here, $\mathcal{L}_{\mathrm{svs}}$ consists of a structure-preserving term $\mathcal{L}_{\mathrm{spv}}$ that aligns pairwise features between teacher and student, and a temporal-consistency term $\mathcal{L}_{\mathrm{temp}}$ that respects spiking dynamics. The mixture weights $\lambda_{\mathrm{spv}}$ and $\lambda_{\mathrm{temp}}$ satisfy $\lambda_{\mathrm{spv}}+\lambda_{\mathrm{temp}}=1$ and can be scheduled from teacher-heavy to task-heavy along training.

\noindent\textbf{Structure-preserving visual loss.}
Given $N$ spatial tokens from the VLM image encoder (teacher) and the student, we form pairwise distance matrices $D^{(t)}, D^{(s)} \in \mathbb{R}^{N\times N}$ in a shared feature space. Concretely, for each token we apply $\mathrm{Norm}(\cdot)$ ($\ell_2$-normalization) and compute cosine distances $D^{(\cdot)}_{ij}=1-\langle \mathbf{f}^{(\cdot)}_i,\mathbf{f}^{(\cdot)}_j\rangle$. The loss we presented in Eq.~\ref{eq:l_vis_snn} minimizes the robust Huber deviation between teacher and student distances, which preserves the global geometry while being less sensitive to outliers than $\ell_2$ with $\mathrm{Huber}(\cdot,\delta)$. This encourages the student to discover teacher-like attention patterns across tokens at very low time steps.
{\setlength\abovedisplayskip{2pt}
\setlength\belowdisplayskip{2pt}
\begin{equation}
\label{eq:l_vis_snn}
\mathcal{L}_{\mathrm{spv}}
= \frac{1}{N^{2}}\sum_{i=1}^{N}\sum_{j=1}^{N}
   \mathrm{Huber}\!\left(D^{(s)}_{ij}-D^{(t)}_{ij},\,\delta\right).
\end{equation}}\\
\noindent\textbf{Temporal consistency.} Let $y_i(t)$ denote the student’s spiking feature for token $i$ at time step $t$, projected by a lightweight adaptor $W_s$ into shared space. Let $u_i$ be the membrane potential feature (time-averaged within a block) and $p_t$ a learnable phase/positional code. Eq.~\ref{eq:l_vis_temp} penalizes disagreement between the normalized projected spike features and their membrane-potential trajectory:
{\setlength\abovedisplayskip{2pt}
\setlength\belowdisplayskip{2pt}
\begin{equation}
\label{eq:l_vis_temp}
\footnotesize
\mathcal{L}_{\mathrm{temp}}
= \frac{1}{NT}\sum_{i=1}^{N}\sum_{t=1}^{T}
   \left\|\,\mathrm{Norm}\!\big(W_s\,y_i(t)\big)
          - \mathrm{Norm}\!\big(u_i+p_t\big)\right\|_2^{2}.
\end{equation}}
This mechanism stabilizes the learning process with discrete spikes and helps the student form smooth, VLM-compatible representations over time.






\vspace{-12pt}
\subsection{Spike Prototype-guided Linguistic Distillation}
\label{sec:lingual_kd}
\noindent \textbf{Insights.} Injecting linguistic supervision from a frozen VLM into a spiking student is difficult for one primary reason, the modalities are misaligned. The student produces time aggregated spike features and class logits, whereas the teacher exposes sentence level text embeddings that live in a cosine normalized space with different scale and geometry. 

Our Spike Prototype-guided Linguistic (SPL) distillation, as depicted in Fig.~\ref{fig:kd_spikformer}, treats the VLM text encoder as a bank of semantic anchors and shapes the student around those anchors at two complementary levels. We first form class text prototypes by prompting the frozen teacher with lightweight templates and aggregating the resulting embeddings. On the student side, we maintain spike prototypes that summarize each class in the student’s latent space, reducing instance noise from discrete spikes. SPL then aligns both the student’s prototypes and its logits to the teacher’s text prototypes through temperature-scaled distributions over the class vocabulary. 


\noindent\textbf{Text embeddings from the frozen VLM.}
Given a class $y\in\{1,\dots,N_{\text{cls}}\}$ and a prompt function $\text{prompt}(\cdot)$ (i.e., ``A photo of a \texttt{[class]}''), the VLM text encoder produces
\begin{equation}
\mathbf{t}^{(k)}_{\text{vlm}} \;=\; f_{\text{txt}}^{\text{(VLM)}}\!\big(\text{prompt}(k)\big), \quad k=1,\dots,N_{\text{cls}},
\label{eq:vlm-text}
\end{equation}
which we adapt to the student's feature dimensionality with a lightweight condensation layer $c_{\text{txt}}(\cdot)$:
{\setlength\abovedisplayskip{2pt}
\setlength\belowdisplayskip{2pt}
\begin{equation}
\tilde{\mathbf{t}}^{(k)} \;=\; c_{\text{txt}}\!\left(\mathbf{t}^{(k)}_{\text{vlm}}\right).
\label{eq:proj-text}
\end{equation}}
Note that all parameters of $f_{\text{txt}}^{\text{(VLM)}}$ remain frozen.

\noindent\textbf{Prototype Formatting.}
For an input image $x$, we extract a frozen teacher image embedding and a student embedding:
\begin{equation}
\mathbf{i}_{\text{vlm}} \;=\; f_{\text{img}}^{\text{(VLM)}}(x), 
\qquad
\mathbf{i}_{s} \;=\; g_{\text{stu}}^{\text{(SF)}}(x),
\end{equation}
where $f_{\text{img}}^{\text{(VLM)}}$ is the VLM visual encoder (frozen) and $g_{\text{stu}}^{\text{(SF)}}$ denotes the pool-averaged spiking token. We then compute per-class cosine similarities between image and text features by element-wise multiplication to form the prototype:
\begin{equation}
\small 
z^{(k)}_{\text{vlm}} \;=\; 
\frac{\mathbf{i}_{\text{vlm}}^{\top}\,\mathbf{t}^{(k)}_{\text{vlm}}}
{\lVert\mathbf{i}_{\text{vlm}}\rVert\,\lVert\mathbf{t}^{(k)}_{\text{vlm}}\rVert},
\qquad
z^{(k)}_{s} \;=\; 
\frac{\mathbf{i}_{s}^{\top}\,\tilde{\mathbf{t}}^{(k)}}
{\lVert\mathbf{i}_{s}\rVert\,\lVert\tilde{\mathbf{t}}^{(k)}\rVert}.
\end{equation}

\noindent\textbf{Linguistic KD objective.}
Let $\mathbf{z}_{\text{vlm}}=[z^{(1)}_{\text{vlm}},\dots,z^{(N_{\text{cls}})}_{\text{vlm}}]$ and 
$\mathbf{z}_{s}=[z^{(1)}_{s},\dots,z^{(N_{\text{cls}})}_{s}]$.
With temperature $T>0$, the linguistic KD loss matches the teacher’s class distribution (over prompts) to the student’s:
\begin{equation}
\mathcal{L}_{\text{spl}}
\;=\;
T^{2}\,\mathrm{KL}\!\left(
\mathrm{softmax}\!\left(\frac{\mathbf{z}_{\text{vlm}}}{T}\right)\,
\Big\|\,
\mathrm{softmax}\!\left(\frac{\mathbf{z}_{s}}{T}\right)
\right).
\label{eq:lingual-kd}
\end{equation}
The $T^{2}$ factor preserves gradient scale under temperature. We backpropagate Eq.~\ref{eq:lingual-kd} only through the student (and $c_{\text{txt}}$), while keeping the VLM parameters frozen. Multiple prompt templates per class can also be averaged in Eq.~\ref{eq:vlm-text} without changing the formulation.
This linguistic objective complements our visual KD term by grounding the student’s spiking features in the VLM’s textual semantics, improving class discrimination especially at low time steps.

\vspace{-10pt}
\subsection{Loss Functions and Optimization}
\label{full_kd}
The total loss for our VL2Spike framework is defined as
\begin{equation}
\label{eq:total-loss}
\mathcal{L}_{\text{total}}
= \lambda_{\text{cls}}\mathcal{L}_{\text{cls}}
+ \lambda_{\text{svs}}\mathcal{L}_{\text{svs}}
+ \lambda_{\text{spl}}\mathcal{L}_{\text{spl}},
\end{equation}



subject to $\lambda_{\text{cls}}+\lambda_{\text{svs}}+\lambda_{\text{spl}}=1$ and $\lambda_{\cdot}\ge0$, where
$\mathcal{L}_{\text{cls}}$ denotes the task classification loss for the Spikformer student,
$\mathcal{L}_{\text{vis}}$ is the visual distillation loss that aligns student visual/spiking features with the frozen VLM's visual encoder,
and $\mathcal{L}_{\text{txt}}$ is the linguistic (semantic) KD loss that aligns the student's mediator representations with the frozen VLM's language embeddings or text-image joint space.
The task classification loss is then defined as:
{\setlength\abovedisplayskip{2pt}
\setlength\belowdisplayskip{2pt}
\begin{equation}
\label{eq:l_cls}
\mathcal{L}_{\text{cls}}
\;=\;
-\sum_{c=1}^{N_{\text{cls}}} y_c \,\log p_s(c\mid x),
\end{equation}}
where $y_c$ is the one-hot indicator for class $c$. For a mini-batch,
we average Eq.~\ref{eq:l_cls} over samples. This objective updates only the \emph{student} including the classifier head, while all VLM parameters remain frozen. The term $\mathcal{L}_{\text{cls}}$ complements our visual and linguistic distillation objectives by directly optimizing task accuracy while frozen VLM provides auxiliary supervision.

\noindent\textbf{Dynamic weighting.} At the start of training we set
$\lambda_{\text{cls}}^{(0)}=0.02$ and
$\lambda_{\text{svs}}^{(0)}=\lambda_{\text{spl}}^{(0)}=0.49$.
We then gradually increase the weight of the classification objective to emphasize specific accuracy as the student stabilizes:
\begin{equation}
\footnotesize
\label{eq:schedule}
\lambda_{\text{cls}}(t)=\lambda_{\text{cls}}^{(0)}+\alpha\frac{t}{T},\qquad
\lambda_{\text{svs}}(t)=\lambda_{\text{spl}}(t)=\frac{1-\lambda_{\text{cls}}(t)}{2},
\end{equation}
where $t$ is the current training step, $T$ is the total number of steps, and $\alpha$ is chosen so that
$\lambda_{\text{cls}}(T)$ reaches the desired final weight. A cosine or piecewise linear warm-up can be used in place of the linear schedule in Eq.~\ref{eq:schedule}.

\noindent\textbf{Optimization.} We backpropagate the gradients of $\mathcal{L}_{\text{total}}$ through the SpikFormer student and the trainable mediator layers only, while keeping all VLM parameters frozen. This preserves the teacher's semantics and ensures that both the visual and linguistic knowledge distilled from the VLM are effectively embedded into the student's spiking parameters.

\vspace{-5pt}
\section{Experiments}

\subsection{Implementation Details and Datasets}

\noindent\textbf{Datasets.} We evaluate the VL2Spike framework on standard static benchmarks CIFAR-10/100 \cite{krizhevsky2009learning}, ImageNet-1K \cite{deng2009imagenet}, two neuromorphic datasets DVS-CIFAR10 \cite{li2017cifar10}, DVS128 Gesture \cite{amir2017low} and two specialized VPR datasets Nordland, Oxford RobotCar~\cite{smith2022openscenevlad, maddern20171}. We follow standard preprocessing and single crop evaluation protocols and omit further details for brevity as these static datasets are well-known.

\noindent\textbf{Implementation Details.} Distillation uses the losses defined in Sec.~\ref{sec:visual_kd} and Sec.~\ref{sec:lingual_kd} and is combined by the overall objective in Eq.~\ref{eq:total-loss}. Logit KD uses temperature $\tau$ (default $\tau{=}2$) and cosine similarity distributions. For linguistic KD we prompt the frozen text encoder with class aware templates, for example \textit{`a photo of a \{class\}'}. Multiple templates per class are averaged to form text prototypes. We train with surrogate gradient backpropagation and rate coding over $T$ time steps with $T{\in}\{4,16\}$ by default, depending on dataset scale. The temporal regularizer in Eq.~\ref{eq:l_vis_temp} uses a learnable phase code $p_t$ and a light adaptor $W_s$. Visual structure alignment employs Huber loss with $\delta{=}1$ on $\ell_2$-normalized token distances. We use PyTorch with AdamW, SpikingJelly, cosine LR decay, and linear warmup for $5\%$ of training.

Default hyperparameters are: learning rate $5\!\times\!10^{-4}$, weight decay $0.05$, gradient clipping at $1.0$, label smoothing $0.1$, and stochastic depth $0.1$ for the student. Data augmentation follows common practices for each dataset. The dynamic weighting for the overall loss as in Eq.~\ref{eq:total-loss} warms in $\lambda_{\text{cls}}$ from $0.02$ to its final value, while $\lambda_{\text{svs}}$ and $\lambda_{\text{spl}}$ decrease from 0.49 proportionally to maintain $\lambda_{\text{cls}}{+}\lambda_{\text{svs}}{+}\lambda_{\text{spl}}{=}1$. The experiments were mainly conducted on systems configured with 4× NVIDIA GeForce RTX 5090 GPUs (32 GB VRAM per GPU).

\noindent \textbf{Teacher and Student Models.} We select the frozen CLIP-style VLM with a ViT-Large backbone \cite{dosovitskiy2020image} as the teacher. For student models, we select lightweight spiking transformer models with a linear classifier head and a single projection or condensation layer for each KD branch (visual or text) to match the dimensionalities of the features.

\vspace{-10pt}
\subsection{Experiments on Conventional Classification Tasks}
\begin{table*}[t!]
\footnotesize
\centering
\setlength{\tabcolsep}{1.3pt}
\caption{Quantitative performance gains of VL2Spike on multiple static datasets, compared against different SNN baselines. Improvements are highlighted to emphasize the impact of the proposed VL2Spike approach. (Top-1 accuracy, \%).}
\label{tab:comparison}
 \resizebox{0.58\linewidth}{!}{\begin{tabular}{l l c c c }
\midrule
\textbf{Model} & \textbf{Distillation} & \textbf{CIFAR-10} \cite{krizhevsky2009learning} & \textbf{CIFAR-100} \cite{krizhevsky2009learning} & \textbf{ImageNet-1K} \cite{deng2009imagenet} \\
\midrule
\midrule
SNN \cite{deng2022temporal}       & \textit{w/o} KD   & 94.50 & 74.72 & 64.79 \\
KDSNN \cite{Xu2023KDSNN}     & \textit{w/ } KD   & 95.20 & 79.06 & 76.44 \\
LaSNN \cite{Hong2023LaSNN}     & \textit{w/ } KD   & 95.79 & 79.99 & 76.38 \\
BKDSNN \cite{Xu2024BKDSNN}    & \textit{w/ } KD   & 96.06 & 81.26 & 77.24 \\
TSSD \cite{zuo2024self}      & \textit{w/ } KD   & 94.41 & 74.69 & 66.13 \\
\midrule
\multirow{3}{*}{Spikformer \cite{zhou2022spikformer}} 
  & \textit{w/o} KD   &  94.76 & 77.62 & 74.81 \\
  & \textit{w/ } VL2Spike   & \textbf{97.20} & \textbf{84.35} & \textbf{81.62}\\
  \rowcolor{yellow!20} & Diff. & +2.44 & +6.73 & +6.81\\
\midrule
    \end{tabular}}
\vspace{-5pt}
\end{table*}


\begin{figure}[t!]
    \centering
    \includegraphics[width=1\linewidth]{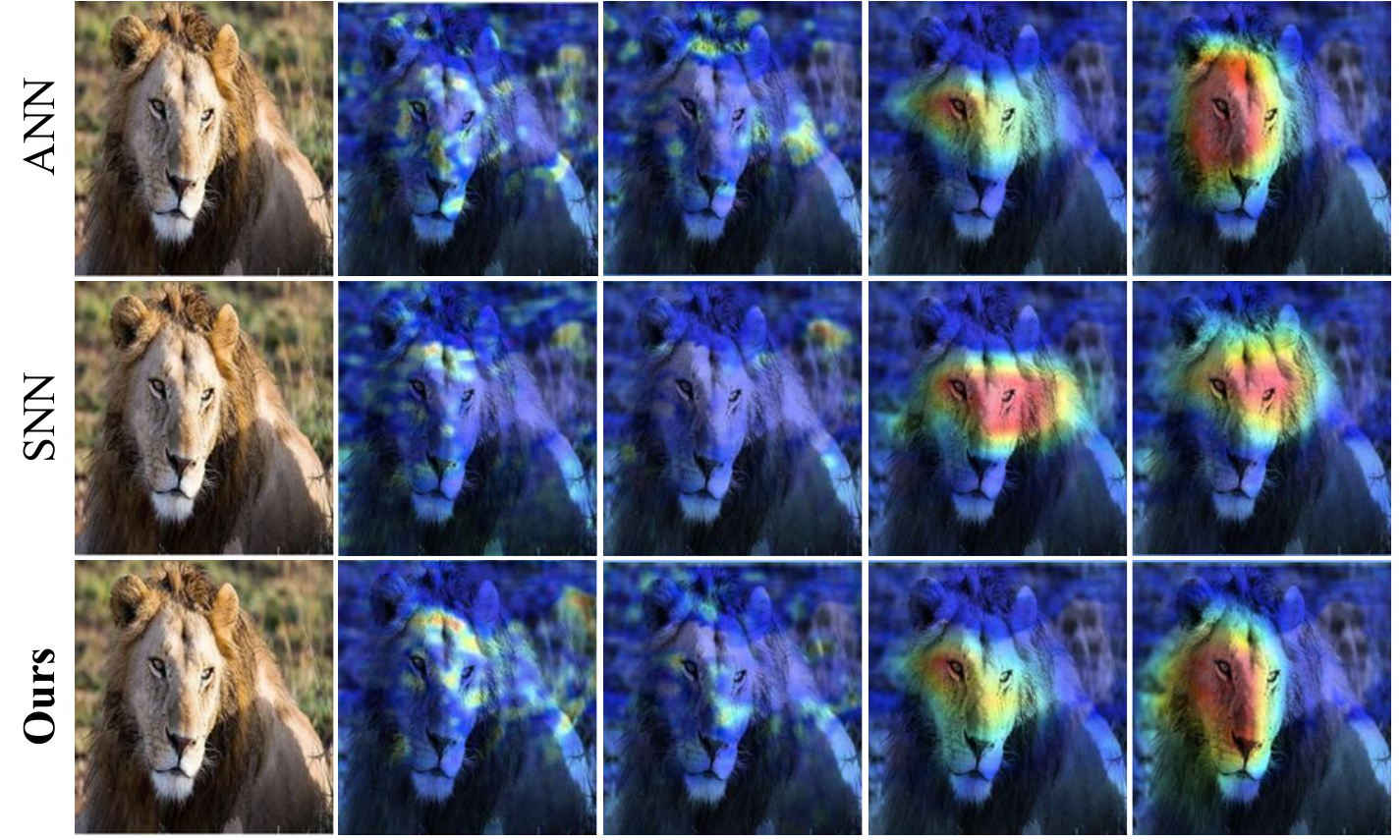}
    \vspace{-18pt}
    \caption{\textbf{An illustration of feature attention for different methods}. Our method yields earlier and progressively sharper focus on the object and fewer background activations, indicating better localization and feature consolidation through depth.}
    \label{fig:attention_map}
    \vspace{-15pt}
\end{figure}

\noindent\textbf{Evaluation of Static Datasets.}
On the static benchmarks (CIFAR-10/100 and ImageNet-1K), Spikformer plus our VL2Spike achieves 97.20 / 84.35 / 81.62 Top-1, setting the best results among SNN methods in Tab.~\ref{tab:comparison}. Relative to the Spikformer baseline without KD, the gains are \textbf{+2.44} on CIFAR-10, \textbf{+6.73} on CIFAR-100, and \textbf{+6.81} on ImageNet-1K, 
{indicating that the benefit grows with dataset complexity and class granularity}.
Compared with BKDSNN, one of the strongest prior KD approaches, VL2Spike still improves by \textbf{+1.14} on CIFAR-10, \textbf{+3.09} on CIFAR-100, and \textbf{+1.69} on ImageNet-1K. Notably, earlier KD methods either underperform on CIFAR-10 compared with a plain SNN or trail the BKDSNN line on CIFAR-100 or ImageNet, suggesting that distilling only logits or spike statistics can be ineffective when teacher capacity is limited or the architecture gap is large. In contrast, VL2Spike consistently boosts the Spikformer student across all three datasets, supporting our claim that combining visual alignment with language-guided class prototypes provides transferable supervision that scales to fine grained and complex large vocabulary recognition.


We also inspect intermediate activations across depth for four models shown as rows in Fig.~\ref{fig:attention_map}. The ANN progressively sharpens attention on the semantically diagnostic regions while the directly trained SNN spreads energy over background fur and sky. A conventional KD baseline reduces some noise but still yields fragmented, off-object highlights. In contrast, ours consistently concentrates on the lion’s facial region from early to late blocks, with tighter, less diffuse responses and minimal background activation. This shows visual-lingual distillation transfers teacher semantics and calibrates spiking features toward object centric evidence.

\noindent\textbf{Evaluation of Neuromorphic Datasets.} We further test the proposed framework on event-based recognition to assess its generality beyond conventional static images. Tab.~\ref{tab:neuro_comparison} summarizes results on DVS-CIFAR10 and DVS128 Gesture. Using the same low latency student, our approach consistently surpasses the non-distilled Spikformer baseline and remains competitive with specialized distillation methods tailored for spiking models. On DVS-CIFAR10, VL2Spike yields an improvement of \textbf{1.94} percentage points over the plain student and edges out prior KD baselines, indicating that the transferred multimodal priors sharpen both spatial selectivity and spike timing on asynchronous streams. 

\begin{table}[t!]
 \captionof{table}{Comparison of VL2Spike framework across neuromorphic datasets (Top-1 accuracy, \%) with different models.}
 \vspace{-5pt}
 \small
    \renewcommand{\arraystretch}{1.15}
    \setlength{\tabcolsep}{5pt}
    \resizebox{\linewidth}{!}{
    \begin{tabular}{lcccccc}
    \toprule
    \multirow{2}{*}{Datasets} & \multicolumn{6}{c}{Methods} \\
    \cmidrule(lr){2-7}
     & SNN & KDSNN & LaSNN & BKDSNN & Spikformer & \textbf{VL2Spike} \\
    \midrule
    DVS-CIFAR10 \cite{li2017cifar10} & 77.33 & 79.90 & 80.10 & 80.85 & 79.20 & \textbf{81.14} \\
    DVS128 Gesture \cite{amir2017low} & 96.87 & - & - & 97.80 & 98.30 & \textbf{98.51} \\
    DVS-Lip \cite{tan2022multi} & - & 51.06 & 50.82 & 52.96 & 49.72 & \textbf{54.35} \\
    N-Caltech101 \cite{orchard2015converting} & - & 69.16 & 69.21 & 69.68 & 68.97 & \textbf{70.03} \\
    \bottomrule
  \end{tabular}
        }
        \label{tab:neuro_comparison}
        \vspace{-5pt}
\end{table}

On DVS128 Gesture, accuracy for modern SNNs is already near saturation. Even in this ceiling regime, VL2Spike still provides a small but reproducible gain over the baseline student. The benefit is achieved without increasing inference steps or adding test time components, which confirms that the additional guidance is fully absorbed into the student’s native spiking dynamics. Taken together, these results show that the proposed distillation is not limited to frame-based imagery. Aligning student spikes to teacher features and language-shaped prototypes improves recognition on neuromorphic data as well, which suggests that the transferred structure acts as a robust inductive bias for event-driven perception. 

\subsection{Experiments on Visual Place Recognition (VPR)}
\begin{figure}[t!]
    \centering
    \includegraphics[width=1\linewidth]{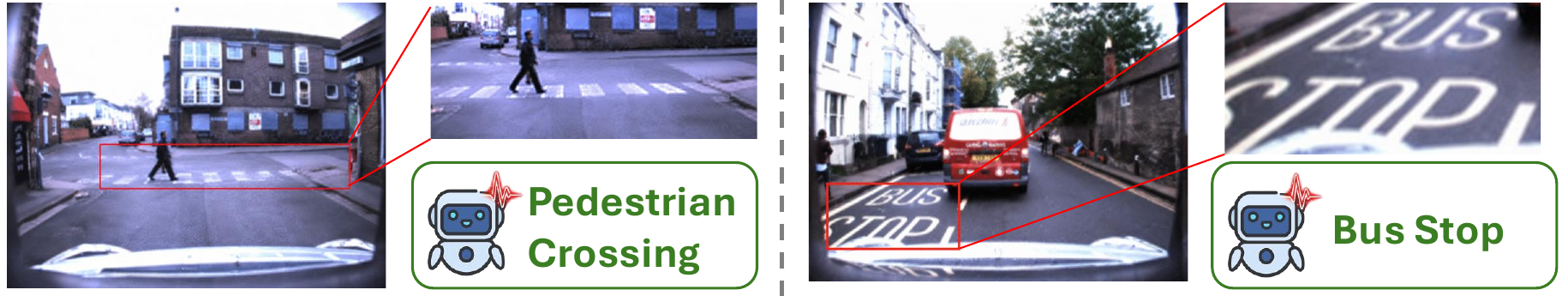}
    \vspace{-15pt}
    \caption{Qualitative examples of VL2Spike on VPR tasks.}
    \label{fig:vpr_results}
    \vspace{-18pt}
\end{figure}

Additionally, the VPR results in Tab.~\ref{tab:vpr_comparison} further demonstrate the generalization ability of VL2Spike beyond conventional classification tasks. 
VL2Spike consistently achieves the best performance on both benchmarks, showing that VLM-guided spike distillation can effectively benefit scene-level place recognition. 
On Nordland, VL2Spike improves the strongest baseline, Spikformer$^*$, from \textbf{53.57\%} to \textbf{57.24\%}, yielding a gain of \textbf{+3.67\%}. 
On the more challenging Oxford RobotCar benchmark, the gain becomes larger, increasing accuracy from \textbf{47.12\%} to \textbf{53.75\%}, with an improvement of \textbf{+6.63\%}. 
This indicates that VL2Spike is especially effective under complex real-world conditions with viewpoint, illumination, and dynamic-scene variations. 
The qualitative examples in Fig.~\ref{fig:vpr_results} further show that VL2Spike can attend to discriminative landmarks, such as pedestrian crossings and bus stops, which are important cues for robust VPR. 
Together, these results demonstrate the effectiveness and transferability of VL2Spike on downstream robotic perception tasks.

\begin{table}[t]
\centering
\caption{Quantitative performance gains of VL2Spike on VPR task. $^*$ denotes necessary task-specific modifications. (Precision at 100\% recall, \%).}
\vspace{-5pt}
\label{tab:precision_100_recall}
\setlength{\tabcolsep}{3pt}
\renewcommand{\arraystretch}{1.05}
\scriptsize
\begin{tabular}{lcc}
\toprule
\textbf{Model} & \textbf{Nordland}~\cite{smith2022openscenevlad} & \textbf{Oxford RobotCar}~\cite{maddern20171} \\
\midrule
SAD & 45.16 & 41.33 \\
NetVLAD & 35.15 & 44.89 \\
Ensemble SNN$^*$ & 52.91 & 43.53 \\
Spikformer$^*$ & 53.57 & 47.12 \\
\rowcolor{yellow!20} VL2Spike$^*$ (ours) & \textbf{57.24} & \textbf{53.75} \\
\bottomrule
\end{tabular}
\label{tab:vpr_comparison}
\vspace{-5pt}
\end{table}

\vspace{-10pt}
\subsection{Ablation and Analysis}
 
\newcommand{\cmark}{\ding{51}}

\begin{table}[t]
\centering
\caption{The efficacy of different loss combinations in our proposed framework (Top-1 accuracy, \%).}
\vspace{-5pt}
\label{tab:loss_ablation}
\renewcommand{\arraystretch}{1.15}
\setlength{\tabcolsep}{4pt}
\resizebox{\linewidth}{!}{
\begin{tabular}{c c c c c c c}
\toprule
\multicolumn{3}{c}{\textbf{Loss Components}} &
\multicolumn{4}{c}{\textbf{Dataset}} \\
\cmidrule(lr){1-3} \cmidrule(lr){4-7}
$\mathcal{L}_{\text{cls}}$ &
$\mathcal{L}_{\text{svs}}$ &
$\mathcal{L}_{\text{spl}}$ &
\textbf{CIFAR-10} &
\textbf{CIFAR-100} &
\textbf{ImageNet-1K} &
\textbf{DVS-CIFAR} \\
\midrule
\cmark &        &        & 95.17 & 78.34 & 77.33 & 78.01 \\
\cmark & \cmark &        & 96.46 & 82.20 & 81.01 & 80.49 \\
\cmark &        & \cmark & 95.72 & 79.93 & 80.38 & 80.05 \\
\rowcolor{yellow!20}
\cmark & \cmark & \cmark &
\textbf{97.20} & \textbf{84.35} & \textbf{81.62} & \textbf{81.14} \\
\bottomrule
\end{tabular}
}
\label{tab:ablation}
\vspace{-6pt}
\end{table}

\noindent\textbf{\textit{Ablation of loss components.}} In our ablation study, as detailed in  Tab.~\ref{tab:loss_ablation}, we studied the effect of various loss functions on the performance of a Spikformer-4-384 model across various datasets for image classification. As listed, we ablate each of the two objectives: spatial-temporal visual spike distillation $\mathcal{L}_{\mathrm{svs}}$, and semantic prototype-guided linguistic distillation $\mathcal{L}_{\mathrm{spl}}$.
Starting from the $\mathcal{L}_{\mathrm{cls}}$ baseline, introducing $\mathcal{L}_{\mathrm{svs}}$ consistently improves performance on all static image datasets indicating that aligning student spike tokens to the teacher’s spatial structure helps preserve salient regions and mitigate resolution limits. Furthermore, the gains from the visual distillation consistently exceed those from the linguistic counterpart, underscoring that visual KD is crucial for exploiting fine grained cues.

\begin{table}[t]
 \captionof{table}{Quantitative performance gains of VL2Spike on static datasets with different students (Top-1 accuracy, \%).}
 \vspace{-5pt}
    \renewcommand{\arraystretch}{1.15}
        \setlength{\tabcolsep}{5pt}
        \resizebox{\linewidth}{!}{
            \begin{tabular}{l l c c c }
\toprule
Model & Method &
\textbf{CIFAR-10} \cite{krizhevsky2009learning} & \textbf{CIFAR-100} \cite{krizhevsky2009learning} & \textbf{ImageNet-1K} \cite{deng2009imagenet} \\
\midrule

\multirow{3}{*}{Spikformer-4-384 \cite{zhou2022spikformer}}
& \textit{w/o} KD 
&  94.67 & 77.62 & 74.81  \\
& \textbf{VL2Spike}
& \textbf{97.20} & \textbf{84.35} & \textbf{81.62}  \\
\rowcolor{yellow!20} & Diff.
& {+2.44} & {+6.73} & {+6.81}  \\
\midrule

\multirow{3}{*}{SEW-ResNet20* \cite{fang2021deep}}
& \textit{w/o} KD 
& 89.07 & 60.16 & 58.63  \\
& \textbf{VL2Spike}
& \textbf{90.42} & \textbf{62.44} & \textbf{61.07}  \\
\rowcolor{yellow!20} & Diff.
& {+1.35} & {+2.28} & {+2.44}  \\
\midrule

\multirow{3}{*}{Spiking-ResNet34 \cite{zheng2021going}}
& \textit{w/o} KD 
& 92.92 & 71.53 & 63.72  \\
& \textbf{VL2Spike}
& \textbf{94.39} & \textbf{74.41} & \textbf{66.19}  \\
\rowcolor{yellow!20} & Diff.
& {+1.47} & {+2.88} & {+2.47} \\
\bottomrule
\end{tabular}
        }
        \label{tab:student_supp}
\end{table}

Using $\mathcal{L}_{\mathrm{spl}}$ instead produces a smaller lift on CIFAR‐style benchmarks but a clearly stronger impact on large vocabulary recognition, where language anchored class prototypes provide semantically structured supervision that reduces confusion among visually similar categories and improves calibration. These results suggest that signals distilled from a VLM’s language encoder are semantically rich and structurally organized, providing strong supervision that effectively guides lightweight students and improves generalization, underscoring the central role of linguistic guidance in the distillation.

Crucially, combining all three losses together yields the best results across all static datasets, outperforming either component alone and evidencing complementary roles. $\mathcal{L}_{\mathrm{svs}}$ sharpens spatial locality and preserves teacher attention, while $\mathcal{L}_{\mathrm{spl}}$ organizes decision boundaries in a semantics aware manifold. The full objective stabilizes optimization and improves generalization at no inference time cost, confirming the effectiveness of our multi-component distillation design. Our integration yields comprehensive distilled knowledge from the teacher by combining visual, semantic, and class related signals.

\noindent\textbf{\textit{Ablation on Students and Teachers.}}
To examine whether the improvement generalizes beyond the Spikformer family, we further evaluate VL2Spike on standard spiking CNN backbones, including SEW-ResNet20 and Spiking-ResNet34. As shown in Tab.~\ref{tab:student_supp}, VL2Spike consistently improves all student architectures across static datasets. 
More importantly, the gains also transfer to convolutional SNNs, improving SEW-ResNet20 by \textbf{+2.28\%} on CIFAR-100 and \textbf{+2.44\%} on ImageNet-1K, and Spiking-ResNet34 by \textbf{+2.88\%} and \textbf{+2.47\%}, respectively. These results indicate that VL2Spike is not architecture-specific, but provides broadly effective supervision for both Transformer-based and CNN-based spiking students.
We also examine three Spikformer variants that differ in width and depth and progressively enable the components of our distillation as shown in Tab.~\ref{tab:student}.
The larger student benefits the most in absolute terms, while the smaller ones still realize meaningful improvements, showing that the method scales with model size without depending on it.

\begin{table}[t!]
    \centering
    \scriptsize 
    
    \begin{minipage}[t]{0.48\linewidth}
        \centering
         \vspace{-3pt}
        \caption{CIFAR-100 results for different architecture of Spiking-Transformer  with selected plug-ins (Top-1 accuracy, \%).}
        \vspace{-5pt}
        \label{tab:student}
        \setlength{\tabcolsep}{2pt} 
        \renewcommand{\arraystretch}{1.2}
        \resizebox{\linewidth}{!}{ 
        \begin{tabular}{l c c c c}
            \toprule
            \textbf{Model (\#Params)} & 
            \rotatebox{45}{\textbf{\textit{w/o} KD}} & 
            \rotatebox{45}{\shortstack{\textbf{+ $\mathcal{L}_{\mathrm{cls}}$}}} & 
            \rotatebox{45}{\shortstack{\textbf{++ $\mathcal{L}_{\mathrm{svs}}$}}} & 
            \rotatebox{45}{\shortstack{\textbf{+++ $\mathcal{L}_{\mathrm{spl}}$}}} \\
            \midrule
            Spik-4-256 (4.15M) & 75.96 & 78.34 & 80.22 & 82.41 \\
            Spik-2-384 (5.76M) & 76.95 & 78.77 & 81.05 & 83.19 \\
            Spik-4-384 (9.32M) & 77.62 & 79.03 & 82.20 & 84.35 \\
            \bottomrule
        \end{tabular}}
    \end{minipage}
    \hfill
    \begin{minipage}[t]{0.48\linewidth}
        \centering
        \vspace{-10pt}
        \caption{Comparison of models based on different teachers. \textit{C-10/100: CIFAR-10/100; IN-1K: ImageNet-1K; DVS: CIFAR-DVS.}}
        \label{tab:teacher}
        \vspace{-5pt}
        \setlength{\tabcolsep}{2pt} 
        \resizebox{\linewidth}{!}{ 
        \begin{tabular}{l l c c c c}
            \toprule
            \textbf{Model} & \textbf{Teacher} & \textbf{C-10} & \textbf{C-100} & \textbf{IN-1K} & \textbf{DVS} \\
            \midrule
            Spikformer & w/o KD & 94.67 & 77.62 & 74.81 & 79.20 \\
            \midrule
            \multirow{5}{*}{VL2Spike} 
            & ViT-S & 96.34 & 82.33 & 79.22 & 80.43 \\
            & ViT-B  & 96.87 & 83.55 & 80.52 & 80.83 \\
            & ViT-L & \textbf{97.20} & \textbf{84.35} & \textbf{81.62} & \textbf{81.14} \\
            & LLaVA & 96.82 & 83.74 & 81.02 & 80.89 \\
            & Qwen2 & 95.98 & 81.83 & 79.67 & 80.21 \\
            \bottomrule
        \end{tabular}}
    \end{minipage}
    
    \vspace{-5pt}
\end{table}


We next study the influence of the teacher as shown in Tab.~\ref{tab:teacher}. Using the same student, we distill from vision language models with increasing backbone capacity, from ViT-Small to ViT-Base and ViT-Large. Performance improves monotonically with stronger teachers on both static and neuromorphic datasets, confirming that richer multimodal priors translate into better visual and spike representations for the student. Even the smallest teacher already surpasses the non-distilled student, while the largest teacher provides the best overall accuracy and stability. The positive trend appears on CIFAR, ImageNet-scale recognition, and DVS benchmarks alike, suggesting that the transferred knowledge is broadly useful rather than dataset specific.

Together, these ablations support our design. The components are complementary, the gains accumulate, and the approach benefits from stronger teachers while remaining effective for SNNs. In practice, a middle-to-large teacher paired with a compact Spikformer offers a favorable trade-off between accuracy and compute, with no extra inference cost. A unified distillation objective that integrates multiple forms of knowledge serves as an inductive bias for event-driven perception, creating a robust mechanism for SNN-KD.

\vspace{-10pt}
\subsection{Efficiency and Energy Consumption Analysis}
As illustrated in Tab.~\ref{tab:energy_cons}, we further confirm the inference-side benefits: on the ResNet-50 setting, our spiking model attains \textbf{72.97\%} accuracy while consuming only \textbf{4.97} mJ, yielding a \textbf{74.67\%} energy reduction compared to the ANN Res-50 (19.63 mJ) and markedly improving accuracy over prior spiking counterparts. Similar trends hold for the Transformer setting, where our approach reaches \textbf{81.62\%} accuracy with \textbf{13.63} mJ (an \textbf{84.30\%} reduction vs.\ the ANN baseline at 86.83 mJ), narrowing the accuracy gap to the teacher while demonstrating the energy advantages.

\begin{table}[t]
  \captionof{table}{Energy consumption on ImageNet.}
  \vspace{-5pt}
    \renewcommand{\arraystretch}{1.15}
        \setlength{\tabcolsep}{2pt}
        \resizebox{\linewidth}{!}{
            \begin{tabular}{l c c c c}
                \toprule
                \textbf{Method} & \textbf{Time} & \textbf{SOPS(G)} ($\downarrow$) & \textbf{Energy Consump-} & \textbf{Top-1} \\
                & \textbf{Step} &  & \textbf{ption (mJ)} ($\downarrow$) & \textbf{Acc} ($\uparrow$) \\
                \midrule
                $Res$-$50$            & 1 & 4.21 (FLOPS)           & 19.63               & 76.43 \\
                Spiking $Res$-$50$    & 4 & 5.43 (28.98\%$\uparrow$) & 5.73 (70.81\%$\downarrow$) & 58.17 (-18.26) \\
                SEW $Res$-$50$        & 4 & 4.96 (17.81\%$\uparrow$) & 5.07 (74.17\%$\downarrow$) & 68.59 (-7.84) \\
                VL2Spike (ours)          & 4 & \textbf{4.77 (11.74\%$\uparrow$)} & \textbf{4.97 (74.67\%$\downarrow$)} & \textbf{72.97 (-3.46)} \\
                \midrule
                $Trans$-$8$-$768$       & 1 & 18.93 (FLOPS)          & 86.83              & 82.37 \\
                $Spik$-$8$-$768$       & 4 & 12.73 (32.78\%$\downarrow$) & 14.23 (83.62\%$\downarrow$) & 76.13 (-6.24) \\
                VL2Spike (ours)          & 4 & \textbf{12.37 (34.65\%$\downarrow$)} & \textbf{13.63 (84.30\%$\downarrow$)} & \textbf{81.62 (-0.75)} \\
                \bottomrule
            \end{tabular}
        }
        \label{tab:energy_cons}
        \vspace{-5pt}
\end{table}

\section{Conclusion and Future Work}
We introduced VL2Spike, a distillation framework that couples a frozen VLM with a spiking student via two complementary losses. Together, these losses turn the VLM into a semantic scaffold while respecting the discrete, time dependent nature of spiking computation. Extensive experiments across static, neuromorphic, and VPR benchmarks on different tasks, multiple Spikformer variants, and extensions to conventional SNNs show that VL2Spike consistently improves accuracy and convergence over traditional KD and non-KD baselines. Our analysis further indicates that combining visual and linguistic distillation yields more structured feature manifolds and better class separation than either signal alone.
\noindent \textbf{Future Work.} We aim to extend our work toward open-ended and temporally complex settings, such as open-vocabulary recognition and continual spike-based learning. We hope VL2Spike encourages further exploration of language-aware spike-based models for perception or reasoning tasks. 


 

\bibliographystyle{IEEEtran}
\bibliography{main}

@String(CVPR  = {IEEE Conf. Comput. Vis. Pattern Recog.})

@String(ECCV  = {Eur. Conf. Comput. Vis.})

@String(ICLR  = {Int. Conf. Learn. Represent.})

@String(AAAI  = {AAAI})

@String(IJCAI = {IJCAI})

@String(CVPR  = {CVPR})

@String(ECCV  = {ECCV})

@String(ICLR  = {ICLR})

@inproceedings{Xu2023KDSNN,
  author    = {Qi Xu and Yaxin Li and Jiangrong Shen and Jian K. Liu and Huajin Tang and Gang Pan},
  title     = {Constructing Deep Spiking Neural Networks from Artificial Neural Networks with Knowledge Distillation},
  booktitle = {Proceedings of the IEEE/CVF Conference on Computer Vision and Pattern Recognition (CVPR)},
  year      = {2023},
}

@article{Hong2023LaSNN,
  author  = {Di Hong and Jiangrong Shen and Yu Qi and Yueming Wang},
  title   = {LaSNN: Layer-wise ANN-to-SNN Distillation for Effective and Efficient Training in Deep Spiking Neural Networks},
  journal = {arXiv preprint arXiv:2304.09101},
  year    = {2023},
  url     = {https://arxiv.org/abs/2304.09101}
}

@inproceedings{Xu2024BKDSNN,
  author    = {Zekai Xu and Kang You and Qinghai Guo and Xiang Wang and Zhezhi He},
  title     = {BKDSNN: Enhancing the Performance of Learning-Based Spiking Neural Networks via Blurred Knowledge Distillation},
  booktitle = {European Conference on Computer Vision (ECCV)},
  year      = {2024},
  publisher = {Springer},
  doi       = {10.1007/978-3-031-72973-7_7}
}

@article{zhou2022spikformer,
  title={Spikformer: When spiking neural network meets transformer},
  author={Zhou, Zhaokun and Zhu, Yuesheng and He, Chao and Wang, Yaowei and Yan, Shuicheng and Tian, Yonghong and Yuan, Li},
  journal={arXiv preprint arXiv:2209.15425},
  year={2022}
}

@inproceedings{Jang2025VL2Lite,
  author    = {Jinseong Jang and Chunfei Ma and Byeongwon Lee},
  title     = {VL2Lite: Task-Specific Knowledge Distillation from Large Vision-Language Models to Lightweight Networks},
  booktitle = {Proceedings of the IEEE/CVF Conference on Computer Vision and Pattern Recognition (CVPR)},
  year      = {2025},
}

@article{neftci2019surrogate,
  title={Surrogate gradient learning in spiking neural networks: Bringing the power of gradient-based optimization to spiking neural networks},
  author={Neftci, Emre O and Mostafa, Hesham and Zenke, Friedemann},
  journal={IEEE Signal Processing Magazine},
  volume={36},
  number={6},
  pages={51--63},
  year={2019},
  publisher={IEEE}
}

@article{hinton2015distilling,
  title={Distilling the knowledge in a neural network},
  author={Hinton, Geoffrey and Vinyals, Oriol and Dean, Jeff},
  journal={arXiv preprint arXiv:1503.02531},
  year={2015}
}

@article{wang2021knowledge,
  title={Knowledge distillation and student-teacher learning for visual intelligence: A review and new outlooks},
  author={Wang, Lin and Yoon, Kuk-Jin},
  journal={IEEE transactions on pattern analysis and machine intelligence},
  volume={44},
  number={6},
  pages={3048--3068},
  year={2021},
  publisher={IEEE}
}

@article{xu2022hierarchical,
  title={Hierarchical spiking-based model for efficient image classification with enhanced feature extraction and encoding},
  author={Xu, Qi and Li, Yaxin and Shen, Jiangrong and Zhang, Pingping and Liu, Jian K and Tang, Huajin and Pan, Gang},
  journal={IEEE Transactions on Neural Networks and Learning Systems},
  volume={35},
  pages={9277--9285},
  year={2022},
  publisher={IEEE}
}

@inproceedings{yu2025fsta,
  title={FSTA-SNN: Frequency-Based Spatial-Temporal Attention Module for Spiking Neural Networks},
  author={Yu, Kairong and Zhang, Tianqing and Wang, Hongwei and Xu, Qi},
  booktitle={Proceedings of the AAAI Conference on Artificial Intelligence},
  volume={39},
  number={21},
  pages={22227--22235},
  year={2025}
}

@article{fang2021deep,
  title={Deep residual learning in spiking neural networks},
  author={Fang, Wei and Yu, Zhaofei and Chen, Yanqi and Huang, Tiejun and Masquelier, Timoth{\'e}e and Tian, Yonghong},
  journal={Advances in Neural Information Processing Systems},
  volume={34},
  pages={21056--21069},
  year={2021}
}

@article{hu2024advancing,
  title={Advancing spiking neural networks toward deep residual learning},
  author={Hu, Yifan and Deng, Lei and Wu, Yujie and Yao, Man and Li, Guoqi},
  journal={IEEE transactions on neural networks and learning systems},
  volume={36},
  number={2},
  pages={2353--2367},
  year={2024},
  publisher={IEEE}
}

@article{zhou2023spikingformer,
  title={Spikingformer: Spike-driven residual learning for transformer-based spiking neural network},
  author={Zhou, Chenlin and Yu, Liutao and Zhou, Zhaokun and Ma, Zhengyu and Zhang, Han and Zhou, Huihui and Tian, Yonghong},
  journal={arXiv:2304.11954},
  year={2023}
}

@article{yao2024spike,
  title={Spike-driven transformer v2: Meta spiking neural network architecture inspiring the design of next-generation neuromorphic chips},
  author={Yao, Man and Hu, JiaKui and Hu, Tianxiang and Xu, Yifan and Zhou, Zhaokun and Tian, Yonghong and Xu, Bo and Li, Guoqi},
  journal={arXiv preprint arXiv:2404.03663},
  year={2024}
}

@article{ghosh2009spiking,
  title={Spiking neural networks},
  author={Ghosh-Dastidar, Samanwoy and Adeli, Hojjat},
  journal={International journal of neural systems},
  volume={19},
  pages={295--308},
  year={2009},
  publisher={World Scientific}
}

@article{maass1997networks,
  title={Networks of spiking neurons: the third generation of neural network models},
  author={Maass, Wolfgang},
  journal={Neural networks},
  volume={10},
  pages={1659--1671},
  year={1997},
  publisher={Elsevier}
}

@inproceedings{abbott2005model,
  title={Model neurons: from hodgkin-huxley to hopfield},
  author={Abbott, Larry F and Kepler, Thomas B},
  booktitle={Statistical Mechanics of Neural Networks: Proceedings of the Xlth Sitges Conference Sitges, Barcelona, Spain, 3--7 June 1990},
  pages={5--18},
  year={2005},
  organization={Springer}
}

@book{gerstner2014neuronal,
  title={Neuronal dynamics: From single neurons to networks and models of cognition},
  author={Gerstner, Wulfram and Kistler, Werner M and Naud, Richard and Paninski, Liam},
  year={2014},
  publisher={Cambridge University Press}
}

@article{diehl2015unsupervised,
  title={Unsupervised learning of digit recognition using spike-timing-dependent plasticity},
  author={Diehl, Peter U and Cook, Matthew},
  journal={Frontiers in computational neuroscience},
  volume={9},
  pages={99},
  year={2015},
  publisher={Frontiers Media SA}
}

@article{bu2023optimal,
  title={Optimal ANN-SNN conversion for high-accuracy and ultra-low-latency spiking neural networks},
  author={Bu, Tong and Fang, Wei and Ding, Jianhao and Dai, PengLin and Yu, Zhaofei and Huang, Tiejun},
  journal={arXiv preprint arXiv:2303.04347},
  year={2023}
}

@inproceedings{wang2022signed,
  title={Signed Neuron with Memory: Towards Simple, Accurate and High-Efficient ANN-SNN Conversion.},
  author={Wang, Yuchen and Zhang, Malu and Chen, Yi and Qu, Hong},
  booktitle={IJCAI},
  pages={2501--2508},
  year={2022}
}

@article{davies2018loihi,
  title={Loihi: A neuromorphic manycore processor with on-chip learning},
  author={Davies, Mike and Srinivasa, Narayan and Lin, Tsung-Han and Chinya, Gautham and Cao, Yongqiang and Choday, Sri Harsha and Dimou, Georgios and Joshi, Prasad and Imam, Nabil and Jain, Shweta and others},
  journal={Ieee Micro},
  volume={38},
  number={1},
  pages={82--99},
  year={2018},
  publisher={IEEE}
}

@article{awais2025foundation,
  title={Foundation models defining a new era in vision: a survey and outlook},
  author={Awais, Muhammad and Naseer, Muzammal and Khan, Salman and Anwer, Rao Muhammad and Cholakkal, Hisham and Shah, Mubarak and Yang, Ming-Hsuan and Khan, Fahad Shahbaz},
  journal={IEEE Transactions on Pattern Analysis and Machine Intelligence},
  year={2025},
  publisher={IEEE}
}

@article{dosovitskiy2020image,
  title={An image is worth 16x16 words: Transformers for image recognition at scale},
  author={Dosovitskiy, Alexey},
  journal={arXiv:2010.11929},
  year={2020}
}

@article{pfeiffer2018deep,
  title={Deep learning with spiking neurons: Opportunities and challenges},
  author={Pfeiffer, Michael and Pfeil, Thomas},
  journal={Frontiers in neuroscience},
  volume={12},
  year={2018},
  publisher={Frontiers}
}

@inproceedings{yang2025efficient,
  title={Efficient ANN-Guided Distillation: Aligning Rate-based Features of Spiking Neural Networks through Hybrid Block-wise Replacement},
  author={Yang, Shu and Yu, Chengting and Liu, Lei and Ma, Hanzhi and Wang, Aili and Li, Erping},
  booktitle={Proceedings of the Computer Vision and Pattern Recognition Conference},
  pages={10025--10035},
  year={2025}
}

@article{yu2025efficient,
  title={Efficient Logit-based Knowledge Distillation of Deep Spiking Neural Networks for Full-Range Timestep Deployment},
  author={Yu, Chengting and Zhao, Xiaochen and Liu, Lei and Yang, Shu and Wang, Gaoang and Li, Erping and Wang, Aili},
  journal={arXiv:2501.15925},
  year={2025}
}

@article{qiu2024self,
  title={Self-architectural knowledge distillation for spiking neural networks},
  author={Qiu, Haonan and Ning, Munan and Song, Zeyin and Fang, Wei and Chen, Yanqi and Sun, Tao and Ma, Zhengyu and Yuan, Li and Tian, Yonghong},
  journal={Neural Networks},
  volume={178},
  pages={106475},
  year={2024},
  publisher={Elsevier}
}

@article{zuo2024self,
  title={Self-Distillation Learning Based on Temporal-Spatial Consistency for Spiking Neural Networks},
  author={Zuo, Lin and Ding, Yongqi and Jing, Mengmeng and Yang, Kunshan and Yu, Yunqian},
  journal={arXiv preprint arXiv:2406.07862},
  year={2024}
}

@article{krizhevsky2009learning,
  title={Learning multiple layers of features from tiny images},
  author={Krizhevsky, Alex and Hinton, Geoffrey and others},
  year={2009},
  publisher={Toronto, ON, Canada}
}

@inproceedings{deng2009imagenet,
  title={Imagenet: A large-scale hierarchical image database},
  author={Deng, Jia and Dong, Wei and Socher, Richard and Li, Li-Jia and Li, Kai and Fei-Fei, Li},
  booktitle={Proceedings of the Computer Vision and Pattern Recognition Conference},
  pages={248--255},
  year={2009},
  organization={Ieee}
}

@article{li2017cifar10,
  title={Cifar10-dvs: an event-stream dataset for object classification},
  author={Li, Hongmin and Liu, Hanchao and Ji, Xiangyang and Li, Guoqi and Shi, Luping},
  journal={Frontiers in neuroscience},
  volume={11},
  pages={244131},
  year={2017},
  publisher={Frontiers}
}

@inproceedings{cherti2023reproducible,
  title={Reproducible scaling laws for contrastive language-image learning},
  author={Cherti, Mehdi and Beaumont, Romain and Wightman, Ross and Wortsman, Mitchell and Ilharco, Gabriel and Gordon, Cade and Schuhmann, Christoph and Schmidt, Ludwig and Jitsev, Jenia},
  booktitle={Proceedings of the IEEE/CVF conference on computer vision and pattern recognition},
  pages={2818--2829},
  year={2023}
}

@article{liu2023visual,
  title={Visual instruction tuning},
  author={Liu, Haotian and Li, Chunyuan and Wu, Qingyang and Lee, Yong Jae},
  journal={Advances in neural information processing systems},
  volume={36},
  pages={34892--34916},
  year={2023}
}

@inproceedings{amir2017low,
  title={A low power, fully event-based gesture recognition system},
  author={Amir, Arnon and Taba, Brian and Berg, David and Melano, Timothy and McKinstry, Jeffrey and Di Nolfo, Carmelo and Nayak, Tapan and Andreopoulos, Alexander and Garreau, Guillaume and Mendoza, Marcela and others},
  booktitle={Proceedings of the IEEE conference on computer vision and pattern recognition},
  pages={7243--7252},
  year={2017}
}

@article{deng2022temporal,
  title={Temporal efficient training of spiking neural network via gradient re-weighting},
  author={Deng, Shikuang and Li, Yuhang and Zhang, Shanghang and Gu, Shi},
  journal={arXiv preprint arXiv:2202.11946},
  year={2022}
}

@article{zhang2024vlm,
  title={Vlm-kd: Knowledge distillation from vlm for long-tail visual recognition},
  author={Zhang, Zaiwei and Meyer, Gregory P and Lu, Zhichao and Shrivastava, Ashish and Ravichandran, Avinash and Wolff, Eric M},
  journal={arXiv preprint arXiv:2408.16930},
  year={2024}
}

@inproceedings{wei2026tpspikformer,
  title={{TP}-Spikformer: Token Pruned Spiking Transformer},
  author={Wei, Wenjie and Zhou, Xiaolong and Zhang, Malu and Belatreche, Ammar and Sun, Qian and Shan, Yimeng and Zhang, Dehao and Zhou, Zijian and Ma, Zeyu and Yang, Yang and Li, Haizhou},
  booktitle={The Fourteenth International Conference on Learning Representations (ICLR)},
  year={2026}
}

@article{wang2025spikcommander,
  title={SpikCommander: A High-performance Spiking Transformer with Multi-view Learning for Efficient Speech Command Recognition},
  author={Wang, Jiaqi and Yu, Liutao and Shen, Xiongri and Guo, Sihang and Zhou, Chenlin and Zhao, Leilei and Zhong, Yi and Zhang, Zhiguo and Ma, Zhengyu},
  journal={arXiv preprint arXiv:2511.07883},
  year={2025}
}

@article{liu2025closer,
  title={A Closer Look at Knowledge Distillation in Spiking Neural Network Training},
  author={Liu, Xu and Xia, Na and Zhou, Jinxing and Xu, Jingyuan and Guo, Dan},
  journal={arXiv preprint arXiv:2511.06902},
  year={2025}
}

@article{orchard2015converting,
  title={Converting static image datasets to spiking neuromorphic datasets using saccades},
  author={Orchard, Garrick and Jayawant, Ajinkya and Cohen, Gregory K and Thakor, Nitish},
  journal={Frontiers in neuroscience},
  volume={9},
  pages={437},
  year={2015},
  publisher={Frontiers Media SA}
}

@inproceedings{tan2022multi,
  title={Multi-grained spatio-temporal features perceived network for event-based lip-reading},
  author={Tan, Ganchao and Wang, Yang and Han, Han and Cao, Yang and Wu, Feng and Zha, Zheng-Jun},
  booktitle={Proceedings of the IEEE/CVF Conference on Computer Vision and Pattern Recognition},
  pages={20094--20103},
  year={2022}
}

@inproceedings{zheng2021going,
  title={Going deeper with directly-trained larger spiking neural networks},
  author={Zheng, Hanle and Wu, Yujie and Deng, Lei and Hu, Yifan and Li, Guoqi},
  booktitle={Proceedings of the AAAI conference on artificial intelligence},
  volume={35},
  number={12},
  pages={11062--11070},
  year={2021}
}

@inproceedings{smith2022openscenevlad,
  title={Openscenevlad: Appearance invariant, open set scene classification},
  author={Smith, William HB and Milford, Michael and McDonald-Maier, Klaus D and Ehsan, Shoaib and Fisher, Robert B},
  booktitle={2022 International Conference on Robotics and Automation (ICRA)},
  pages={4578--4584},
  year={2022},
  organization={IEEE}
}

@article{maddern20171,
  title={1 year, 1000 km: The oxford robotcar dataset},
  author={Maddern, Will and Pascoe, Geoffrey and Linegar, Chris and Newman, Paul},
  journal={The International Journal of Robotics Research},
  volume={36},
  number={1},
  pages={3--15},
  year={2017},
  publisher={SAGE Publications Sage UK: London, England}
}


\newpage

 




\vfill

\end{document}